\documentclass{article}
\usepackage{spconf,amsmath,graphicx,hyperref}
\usepackage{amsmath,amssymb}
\usepackage{booktabs}
\usepackage{pifont}
\usepackage[table]{xcolor}

\title{Vision KAN: Towards an Attention-Free Backbone for Vision with Kolmogorov-Arnold Networks}
\name{Zhuoqin Yang$^{1,2}$, Jiansong Zhang$^{1}$, Xiaoling Luo$^{1}$, Xu Wu$^{1}$, Zheng Lu$^{2,*}$, Linlin Shen$^{3,*}$\thanks{*Corresponding authors} \thanks{© 2026 IEEE. Personal use of this material is permitted. Permission from IEEE must be obtained for all other uses, in any current or future media, including reprinting/republishing this material for advertising or promotional purposes, creating new collective works, for resale or redistribution to servers or lists, or reuse of any copyrighted component of this work in other works.}}
\address{
    $^{1}$College of Computer Science and Software Engineering, Shenzhen University, Shenzhen, China \\
    $^{2}$School of Computer Science, University of Nottingham Ningbo China, Ningbo, China  \\
    $^{3}$School of Artificial Intelligence, Shenzhen University, Shenzhen, China \\
}

\begin{document}
% \ninept
%
\maketitle
\begin{abstract}

Attention mechanisms have become a key module in modern vision backbones due to their ability to model long-range dependencies. However, their quadratic complexity in sequence length and the difficulty of interpreting attention weights limit both scalability and clarity. Recent attention-free architectures demonstrate that strong performance can be achieved without pairwise attention, motivating the search for alternatives. In this work, we introduce Vision KAN (ViK), an attention-free backbone inspired by the Kolmogorov-Arnold Networks. At its core lies MultiPatch-RBFKAN, a unified token mixer that combines (a) patch-wise nonlinear transform with Radial Basis Function-based KANs, (b) axis-wise separable mixing for efficient local propagation, and (c) low-rank global mapping for long-range interaction. Employing as a drop-in replacement for attention modules, this formulation tackles the prohibitive cost of full KANs on high-resolution features by adopting a patch-wise grouping strategy with lightweight operators to restore cross-patch dependencies. Experiments on ImageNet-1K show that ViK achieves competitive accuracy with linear complexity, demonstrating the potential of KAN-based token mixing as an efficient and theoretically grounded alternative to attention. Code: \url{https://github.com/SeriYann/ViK}

\end{abstract}

\section{Introduction}

Vision backbones in recent years have been dominated by attention-based architectures.  
The Vision Transformer (ViT) \cite{dosovitskiy2020image} demonstrated that modeling images as sequences of patches with self-attention can rival or surpass convolutional networks, establishing attention as a general-purpose visual modeling tool. Subsequent variants further extended this line of work, such as Data-efficient Image Transformer (DeiT) \cite{touvron2021training}, which improved training efficiency and practicality, while Pyramid Vision Transformer (PVT) \cite{wang2021pyramid} introduced hierarchical feature maps with spatial reduction, bridging Transformers with multi-scale representations.

Despite their success, attention modules face two major limitations: first, their computational and memory complexity grows quadratically with the number of tokens, making them expensive to scale to high-resolution images \cite{gu2023mamba}; second, the token-to-token affinity matrix in self-attention is not directly linked to semantic or structural cues, limiting interpretability \cite{ bastings2020elephant}. These limitations have motivated the development of attention-free backbones that retain the Transformer architecture while replacing attention with alternative token mixers. MLP-Mixer \cite{tolstikhin2021mlp} demonstrated that channel and token MLPs can achieve competitive performance, while MetaFormer \cite{yu2022metaformer} further established that the backbone scaffold, consisting of normalization, token mixers, and channel-wise MLPs, can be effective even without attention mechanisms. Together, these works indicate that strong performance in vision does not inherently require quadratic attention, motivating the search for principled and efficient function-based alternatives.

\begin{figure*}[h]
    \centering
    \includegraphics[width=0.75\textwidth]{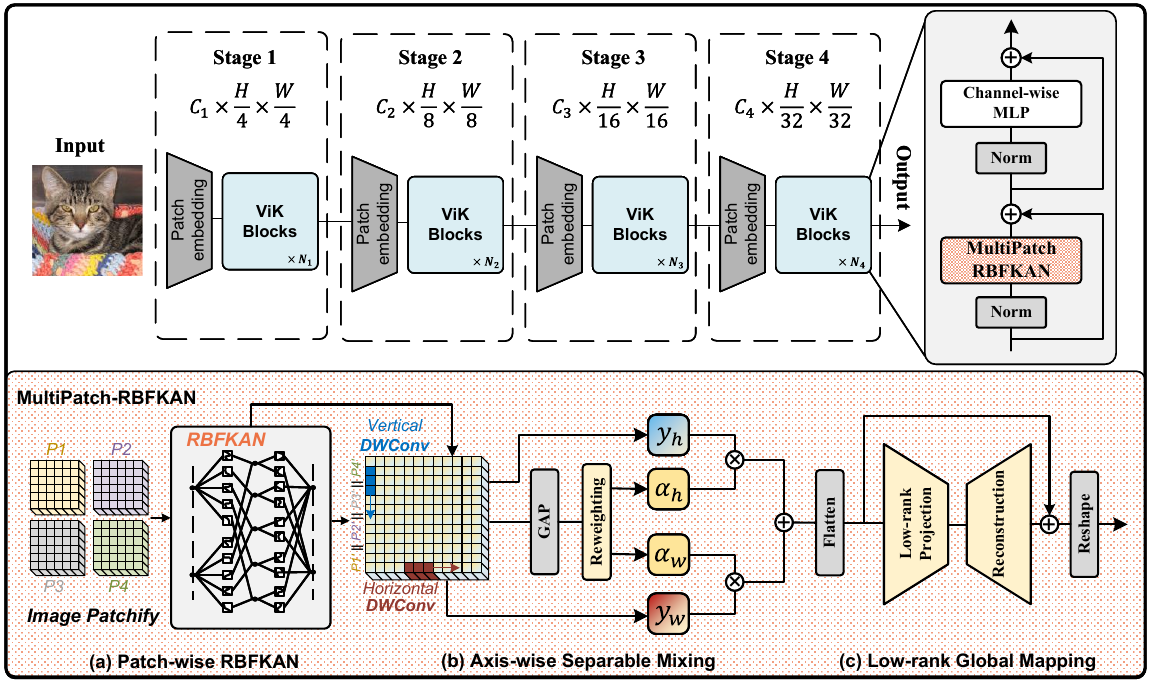}

    \caption{Overview of the proposed Vision KAN (ViK). The backbone adopts a hierarchical design with four stages, where feature maps are progressively downsampled and processed by ViK blocks. Each ViK block contains the MultiPatch-RBFKAN module, which integrates (a) patch-wise nonlinear modeling with RBFKAN, (b) axis-wise separable depthwise convolutions for direction-sensitive local mixing, and (c) a low-rank global path for efficient long-range dependency modeling. Here, $\oplus$ and $\otimes$ denotes element-wise addition and multiplication, while ‖ denotes concatenation.}
    \label{fig:overview}
\end{figure*}

In parallel, the Kolmogorov-Arnold representation theorem (K-A theorem) \cite{kolmogorov1961representation} offers a powerful theoretical insight: any multivariate continuous function can be represented as compositions of univariate functions. This motivates a function-based perspective for modeling token interactions, in contrast to pairwise similarity or linear projection. Recent advances in Kolmogorov-Arnold Networks (KANs) \cite{liu2024kan, liu2024kan2} and their applications in diverse areas \cite{yang2024activation, yu2025exploring, yang2025medkan} have shown strong approximation ability, local interpretability, and adaptability across domains, suggesting a promising direction for designing new backbone modules.

Building on these insights, we introduce \textbf{Vision KAN (ViK)}, an attention-free backbone that replaces pairwise attention with a function-based token mixer inspired by the K–A theorem. The central component is MultiPatch-RBFKAN, which integrates patch-wise nonlinear modeling with RBF bases, axis-wise separable mixing, and low-rank global mapping. A key challenge is that directly applying full KAN layers to high-resolution features is computationally prohibitive. We therefore adopt a patch-wise grouping strategy to make KAN feasible at scale. To compensate for the limited cross-patch interaction introduced by grouping, separable mixing enhances local propagation while the global mapping restores long-range dependencies. This unified design balances expressivity and efficiency. Experiments on ImageNet-1K show that ViK achieves competitive accuracy with linear complexity and a principled theoretical foundation.

The key contributions of this work are:
\begin{itemize}
    \item We propose \textbf{Vision KAN (ViK)}, an attention-free vision backbone that replaces self-attention with a KAN-inspired token mixer,  offering an efficient alternative to the attention module.  

    \item We design \textbf{MultiPatch-RBFKAN}, a unified block that combines patch-wise nonlinear RBF transform, axis-wise separable mixing, and low-rank global mapping to balance expressivity and efficiency.  

    \item Experiments on ImageNet-1K show that ViK achieves competitive accuracy with linear complexity, validating the effectiveness of KAN-based token mixing in large-scale vision tasks.
\end{itemize}

\section{Method}

\subsection{Preliminaries: Kolmogorov-Arnold Networks}
The Kolmogorov-Arnold representation theorem states that any multivariate continuous function can be expressed as compositions of univariate functions, i.e., $f(x_1,\dots,x_n)=\sum_q \phi_q\!\left(\sum_p \phi_{pq}(x_p)\right)$, 
where $\phi(\cdot)$ are continuous univariate functions. 
This suggests that complex mappings can be approximated without explicitly modeling pairwise interactions, 
in contrast to self-attention.

Kolmogorov-Arnold Networks (KANs) implement this principle by replacing fixed activations with parametric basis expansions:
\begin{equation}
\label{eqn:Bx}
\phi(x) = \sum_{j=1}^{M} w_j \cdot B_j(x),
\end{equation}
where $B_j(\cdot)$ are basis functions, $w_j$ are trainable weights, and $M$ is the number of basis functions. 
Compared to conventional MLPs, KANs provide stronger function approximation and local interpretability, since each basis function can be explicitly visualized. 
In this work, we exploit this property to design an \textbf{attention-free vision backbone} that avoids quadratic self-attention yet retains high representational power.

% \subsection{Preliminaries: Kolmogorov-Arnold Networks}
% According to the Kolmogorov--Arnold representation theorem, 
% any multivariate continuous function can be represented as a composition of univariate functions, 
% i.e., $f(x_1,\dots,x_n)=\sum_q \phi_q\!\left(\sum_p \phi_{pq}(x_p)\right)$.
% This suggests that complex mappings can be approximated without explicitly modeling pairwise interactions, 
% in contrast to self-attention.

% Kolmogorov--Arnold Networks (KANs) operationalize this idea by replacing fixed activations with parametric basis expansions, 
% $\phi(x)=\sum_{j=1}^{M} w_j B_j(x)$, 
% where $B_j(\cdot)$ are learnable basis functions.
% Compared to conventional MLPs, KANs provide stronger nonlinear approximation with a compact functional form, 
% making them a natural building block for attention-free vision backbones.

\subsection{Overall Architecture}
\textbf{ViK} is an attention-free vision backbone with a hierarchical design comprising: a convolutional patch embedding, four stages of ViK blocks with decreasing spatial resolution and increasing channels, and a final normalization and classification head.
Each ViK block replaces self-attention with our \textbf{MultiPatch-RBFKAN}, a unified token mixer that achieves local--to--global interactions with linear complexity in the number of tokens, serving as a drop-in alternative to quadratic attention.

\subsection{KAN-based Token Mixer}
Applying a global KAN directly over all tokens is computationally prohibitive. 
Instead, ViK adopts a structured replacement: \textbf{MultiPatch-RBFKAN}, which integrates three complementary operators into a single token-mixing module:
(a) a patch-wise KAN for nonlinear local modeling,
(b) axis-wise separable mixing to exchange information across patches at linear cost, and
(c) a low-rank global mapping to capture long-range dependencies without quadratic overhead.

\textbf{(a) Patch-wise RBFKAN.}  
Given an input feature map, we first divide it into non-overlapping $p \times p$ patches. 
Following Eq.~\ref{eqn:Bx}, where $B(x)$ denotes the basis functions in KAN, we implement $B(x)$ with Radial Basis Functions (RBFs) to model local nonlinear interactions within each patch. 
Concretely, each patch vector is transformed as:
\begin{equation}
\phi(x) = \sum_{j=1}^{M} w_j \cdot \exp\!\left(-\frac{\|x - \mu_j\|^2}{2\sigma_j^2}\right),
\end{equation}
where $\mu_j$ and $\sigma_j$ are learnable centers and widths, and $w_j$ are trainable scaling weights.  
Unlike the B-spline basis commonly used in KANs, which requires recursive computation, RBFs allow all basis activations to be calculated in parallel, making them more GPU-friendly and efficient in large-scale vision tasks~\cite{li2024kolmogorovarnold}.

\textbf{(b) Axis-wise Separable Mixing.} 
While patch-wise KAN captures intra-patch dependencies, modeling cross-patch interactions is equally essential.  
To achieve this efficiently, we apply two depthwise convolutions (DW) along horizontal and vertical axes, followed by global average pooling (GAP) and an MLP-based reweighting:

\begin{equation}
\begin{aligned}
&\hat{y} = \alpha_h \cdot DW_h(y) + \alpha_w \cdot DW_w(y), \\
&[\alpha_h, \alpha_w] = \text{Softmax}\!\left(f_{\text{MLP}}(\text{GAP}(y))\right).
\end{aligned}
\end{equation}
This mechanism introduces direction-sensitive local mixing, enabling the model to adaptively emphasize horizontal or vertical dependencies depending on image structures.

\textbf{(c) Low-rank Global Mapping.}  
Patch-wise local operators are insufficient to capture long-range dependencies.  
To introduce global context efficiently, we reshape each channel into a length-$N$ token vector ($N=H \times W$) and apply a low-rank projection:
\begin{equation}
y_{\text{global}} = \mathbf{Q}\mathbf{P} y, 
\quad \mathbf{P} \in \mathbb{R}^{r \times N}, \;
\mathbf{Q} \in \mathbb{R}^{N \times r}, \;
r \ll N.
\end{equation}
Here $\mathbf{P}$ compresses spatial tokens into a rank-$r$ latent space, and $\mathbf{Q}$ projects them back, enabling efficient global information exchange.

% \subsection{Complexity Analysis}
% The computational cost of each component is as follows:
% \begin{itemize}
%     \item Patch-wise RBFKAN: $O(B\cdot C\cdot H\cdot W \cdot G)$, where $G$ is the number of basis functions;
%     \item H/W Separable Mixing: $O(B\cdot C\cdot H\cdot W \cdot k)$, with kernel size $k$;
%     \item Low-rank Global Linear: $O(B\cdot C\cdot H\cdot W \cdot r)$, where $r \ll N$.
% \end{itemize}
% Together, these three modules provide complementary benefits: 
% patch-wise KAN enhances nonlinear expressivity, separable mixing strengthens local connectivity, and low-rank mapping supplies long-range global context. 
% Overall, ViK achieves \textbf{linear complexity in image size}, while mimicking the expressive power of attention in a more efficient and interpretable way.

% \subsection{Complexity Analysis}
% Let $N{=}H\!\times\!W$, patch size $p$, feature dimension $C$, number of basis $M$, and global rank $r\!\ll\!N$.  
% The patch-wise RBFKAN costs $\mathcal{O}(NMF)$, where $F=p^2$ is fixed.  
% Separable propagation costs $\mathcal{O}(N C k)$.  
% The low-rank global path costs $\mathcal{O}(Nr)$.  
% Therefore each block scales linearly:
% \[
% \mathcal{O}\big(N \cdot (MF + Ck + r)\big),
% \]
% which is significantly more efficient than $\mathcal{O}(N^2 C)$ self-attention in both time and memory.

\subsection{Complexity Analysis}
Let $N = H \times W$ be the spatial resolution, $p$ the patch size, $C$ the channel dimension, $M$ the number of basis functions, and $r \ll N$ the global rank.  
The computational cost of each component in MultiPatch-RBFKAN is:
\begin{itemize}
    \item \textbf{Patch-wise RBFKAN:} $\mathcal{O}(N \cdot C \cdot M \cdot F)$, where $F = p^2$ is the patch dimension;
    \item \textbf{Axis-wise Separable Mixing:} $\mathcal{O}(N \cdot C \cdot k)$, with kernel size $k$;
    \item \textbf{Low-rank Global Mapping:} $\mathcal{O}(N \cdot C \cdot r)$.
\end{itemize}
Therefore, the overall complexity per block is
\[
\mathcal{O}\big(N \cdot C \cdot (M F + k + r)\big),
\]
which scales linearly with image size. Compared with quadratic self-attention $\mathcal{O}(N^2 \cdot C)$, this yields a more efficient attention-free alternative while preserving expressive capacity.

\section{Experiments and Results}

\subsection{Experimental Setup}
We evaluate ViK on the ImageNet-1K \cite{deng2009imagenet} classification benchmark (1.28M training images, 50K validation images, 1K classes). 
All models are trained from scratch for 300 epochs using AdamW \cite{loshchilov2017decoupled} optimizer with weight decay of 0.05 and a peak learning rate of $1\times10^{-3}$. 
We adopt the commonly used augmentation and optimization strategy in modern vision backbones, ensuring a fair comparison \cite{wang2021pyramid}.
We report Top-1 accuracy, number of parameters, and GFLOPs measured at $224 \times 224$ input resolution. 
All experiments are conducted on 4 $\times$ NVIDIA RTX A6000 GPUs.

\subsection{Comparison with Representative Vision Backbones}
% Table~\ref{tab:imagenet} compares ViK with a set of widely adopted and representative vision backbone models, including CNNs (ResNet \cite{he2016deep}), Transformers (ViT \cite{dosovitskiy2020image}, DeiT \cite{touvron2021training}, PVT \cite{wang2021pyramid}), and attention-free architectures (ResMLP \cite{touvron2022resmlp}).  
% Our ViK-Small achieves 76.5\% Top-1 accuracy with only 1.6 GFLOPs, outperforming early Transformers like ViT-Ti and DeiT-Ti by a large margin, with slightly higher computational cost.  
% Compared to ResMLP-S12 and PVT-Tiny, ViK-Small achieves comparable or superior accuracy with fewer FLOPs and parameters.

% At the larger scale, ViK-Base obtains 80.3\% Top-1 accuracy with 3.2 GFLOPs, surpassing ResNet-50 and matching or outperforming attention-based and attention-free backbones — all with significantly reduced complexity.
% This demonstrates that ViK achieves strong accuracy-efficiency trade-offs, validating its lightweight design and effective token mixing strategy without relying on attention mechanisms.

Table~\ref{tab:imagenet} compares ViK with a set of widely adopted vision backbone models, including CNN (ResNet \cite{he2016deep}), Transformers (ViT \cite{dosovitskiy2020image}, DeiT \cite{touvron2021training}, PVT \cite{wang2021pyramid}), and MLP-like architecture (ResMLP \cite{touvron2022resmlp}).  
Our ViK-Small achieves 76.5\% Top-1 accuracy with only 1.6 GFLOPs, providing clear improvements over early Transformers such as ViT-Ti and DeiT-Ti, with only slightly higher computational cost.  
Compared to ResMLP-S12 and PVT-Tiny, ViK-Small attains similar or higher accuracy while requiring fewer FLOPs and parameters.

At a larger scale, ViK-Base reaches 80.3\% Top-1 accuracy with 3.2 GFLOPs, surpassing ResNet-50 and matching or exceeding the performance of DeiT-Small, PVT-Small, and ResMLP-S24, while maintaining substantially lower complexity.  
These results demonstrate that ViK achieves a favorable balance between accuracy and efficiency, validating its lightweight design and effective token mixing strategy without relying on attention mechanisms.

\begin{table}[h]
\centering
\caption{Comparison between architectures on ImageNet-1K classification.}
\resizebox{0.48\textwidth}{!}{ 
\label{tab:imagenet}
\begin{tabular}{lcccc}
\toprule
Model & Type & Params (M) & GFLOPs & Top-1 (\%) \\
\midrule
ResNet-18 \cite{he2016deep} & CNN & 11.7 & 1.8 & 70.6 \\
ViT-Ti/16 \cite{dosovitskiy2020image} & Attention & 5.7 & 1.3 & 72.7 \\
DeiT-Tiny \cite{touvron2021training} & Attention  & 5.7 & 1.3 & 72.2 \\
PVT-Tiny \cite{wang2021pyramid} & Attention  & 13.2 & 1.9 & 75.1 \\
ResMLP-S12 \cite{touvron2022resmlp}  & MLP  & 15 & 3.0 & 76.6 \\
\rowcolor{gray!15}
ViK-Small (ours) & KAN & 13.5 & 1.6 & 76.5 \\
\midrule
ResNet-50 \cite{he2016deep} & CNN & 25.6 & 4.1 & 79.2 \\
ViT-S/16 \cite{dosovitskiy2020image} & Attention & 22.1 & 4.6 & 78.8 \\
DeiT-Small \cite{touvron2021training} & Attention & 22.1 & 4.6 & 79.8 \\
PVT-Small \cite{wang2021pyramid} & Attention  & 24.5 & 3.8 & 79.8 \\
ResMLP-S24 \cite{touvron2022resmlp}  & MLP  & 30 & 6.0 & 79.4 \\
\rowcolor{gray!15}
ViK-Base (ours) & KAN & 24.9 & 3.2 & 80.3 \\
\bottomrule
\end{tabular}
}
\end{table}

\vspace{-3mm}

\begin{table}[h]
\caption{Ablation study on ViK-Small. “Activation” denotes the type of basis functions in KAN (RBF, B-spline, Wavelet, or replaced by an MLP). “\#Basis” indicates the number of basis functions. }
\centering
\resizebox{0.48\textwidth}{!}{ 

\label{tab:ablation}
\begin{tabular}{lccccc}
\toprule
Activation & \#Basis & Separable Mixing & Global Mapping & Top-1 (\%) \\
\midrule
RBF  & 4 & \checkmark & \checkmark & 74.8 \\
RBF  & 6 & \checkmark & \checkmark & 75.7 \\
RBF  & 10 & \checkmark & \checkmark & 76.4 \\
\midrule
B-spline & 8 & \checkmark & \checkmark & 73.8 \\
Wavelet  & 8 & \checkmark & \checkmark & 75.2 \\
\midrule
MLP  & - & \checkmark & \checkmark  & 72.1 \\
RBF  & 8 & \checkmark & \ding{55} &  74.6 \\
RBF  & 8 & \ding{55} & \checkmark & 73.9 \\
\midrule
\rowcolor{gray!15}
RBF  & 8 & \checkmark & \checkmark & \textbf{76.5} \\
\bottomrule
\end{tabular}
}
\end{table}
\vspace{-3mm}

\subsection{Ablation Study}
We conduct ablation experiments on ViK-Small to examine the effect of activation functions, the number of basis functions, and the structural components of MultiPatch-RBFKAN.  
Table~\ref{tab:ablation} summarizes the results.

\textbf{Number of basis functions.}  
Increasing the number of RBFs steadily improves accuracy (from 74.8\% with 4 bases to 76.4\% with 10 bases). Performance peaks at $M{=}8$ (76.5\%) and slightly decreases at $M{=}10$ (76.4\%), indicating saturation with mild overfitting or optimization noise.  
Given that the computational cost grows linearly with $M$, we adopt $M{=}8$ as the default trade-off between accuracy and efficiency.

\textbf{Activation functions.}  
Replacing the basis functions in KAN with different alternatives shows that RBF consistently outperforms B-spline and wavelet bases. Moreover, using comparable-size MLP in place of KAN leads to a significant drop (${-}4.4\%$). This supports our design choice: RBFs are more GPU-friendly and provide stronger nonlinear approximation in vision tasks.

\textbf{Separable mixing and global mapping.} 
Removing either axis-wise separable mixing or the low-rank global mapping leads to a clear performance drop (from 76.5\% to 74.6\% and 73.9\%, respectively). 
Together with the MLP replacement result, these findings validate that our proposed techniques make KAN more suitable for vision tasks by complementing local nonlinear modeling with axis-wise mixing and global interaction.

% \subsection{Ablation Study}
% We conduct ablation experiments on ViK-Small to evaluate the effects of the basis functions, the number of bases, and key architectural components in MultiPatch-RBFKAN. 
% As shown in Table~\ref{tab:ablation}, increasing the number of RBF bases improves accuracy up to $M{=}8$, beyond which performance saturates, leading us to adopt $M{=}8$ as a balanced choice.
% RBF bases consistently outperform B-spline, wavelet, and MLP alternatives, while removing either axis-wise separable mixing or low-rank global mapping results in noticeable degradation.
% These results confirm that both the RBF-based nonlinear modeling and the proposed structural components are critical to ViK’s performance.

\subsection{Visualization of Learned RBF Mappings}

To gain insights into the internal workings of ViK, we visualize the univariate functions $\phi(x)$ learned by RBF across different stages and blocks. Each curve corresponds to a specific input dimension within a patch. The results show that RBF effectively captures nonlinear transformations: functions in shallow stages exhibit more oscillatory shapes sensitive to local variations, while those in deeper stages converge to smoother forms, indicating progressive abstraction toward stable, discriminative features. These univariate mappings provide a clear form of local interpretability.

\begin{figure}[h]
    \centering
    \includegraphics[width=0.48\textwidth]{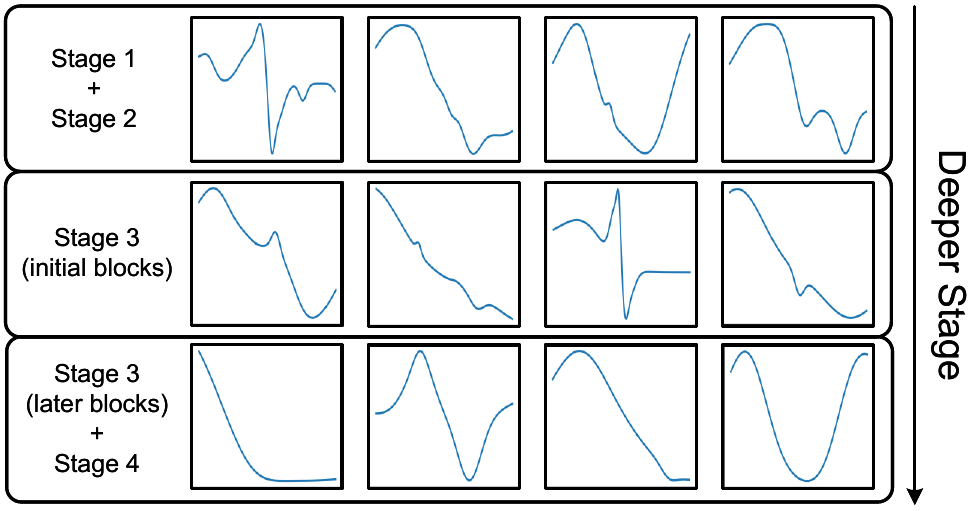}
    % \caption{Each curve shows the univariate function learned by RBFs. Shallow stages exhibit oscillatory nonlinearities, while deeper stages converge to smoother mappings.}
    \caption{Examples of the univariate functions $\phi(x)$ learned by RBF across different stages of ViK. Shallow stages exhibit oscillatory nonlinearities, while deeper stages converge to smoother mappings, indicating progressive abstraction.}
    \label{fig:visualization}
\end{figure}

\section{Conclusion}

We proposed \textbf{ViK}, an attention-free vision backbone that replaces quadratic-cost self-attention with a function-based token mixer inspired by the Kolmogorov-Arnold representation theorem. The proposed \textbf{MultiPatch-RBFKAN} integrates patch-wise nonlinear RBF expansions, axis-wise separable mixing, and low-rank global mapping into a unified module, enabling efficient local and global interactions. Experiments on ImageNet-1K demonstrate that ViK achieves competitive accuracy compared with classical attention-based models, while maintaining linear complexity with respect to image size. These results highlight the potential of KAN–based nonlinear transformations as a principled foundation for vision backbones, suggesting a new direction for building efficient architectures in computer vision. 

% % -------------------------------------------------------------------------
\section{Acknowledgment}
% This work was supported by the National Natural Science Foundation of China (62576216, 62502320), Guangdong Provincial Key Laboratory (2023B1212060076), Natural Science Foundation of Guangdong Province (2025A1515010184), Shenzhen Science and Technology Innovation Committee (JCYJ20240813141424032), and the Foundation for Young Innovative Talents in Ordinary Universities of Guangdong (2024KQNCX042).
This work was supported by the National Natural Science Foundation of China under Grant 62576216 and 62502320, the Guangdong Provincial Key Laboratory under Grant 2023B1212060076, the Natural Science Foundation of Guangdong Province under Grant 2025A1515010184, the project of Shenzhen Science and Technology Innovation Committee under Grant JCYJ20240813141424032, and the Foundation for Young Innovative Talents in Ordinary Universities of Guangdong under Grant 2024KQNCX042.

\bibliographystyle{IEEEbib}
\bibliography{refs}

@article{gu2023mamba,
	author = {Gu, Albert and Dao, Tri},
	journal = {arXiv preprint arXiv:2312.00752},
	title = {Mamba: Linear-time sequence modeling with selective state spaces},
	year = {2023}}

@article{tolstikhin2021mlp,
	author = {Tolstikhin, Ilya O and Houlsby, Neil and Kolesnikov, Alexander and Beyer, Lucas and Zhai, Xiaohua and Unterthiner, Thomas and Yung, Jessica and Steiner, Andreas and Keysers, Daniel and Uszkoreit, Jakob and others},
	journal = {Advances in neural information processing systems},
	pages = {24261--24272},
	title = {Mlp-mixer: An all-mlp architecture for vision},
	volume = {34},
	year = {2021}}

@article{li2024kolmogorovarnold,
  title={Kolmogorov-arnold networks are radial basis function networks},
  author={Li, Ziyao},
  journal={arXiv preprint arXiv:2405.06721},
  year={2024}
}

@inproceedings{liu2024kan,
  title={KAN: Kolmogorov--Arnold Networks},
  author={Liu, Ziming and Wang, Yixuan and Vaidya, Sachin and Ruehle, Fabian and Halverson, James and Soljacic, Marin and Hou, Thomas Y and Tegmark, Max},
  booktitle={International conference on learning representations}, 
  year = {2025}
}

@inproceedings{touvron2021training,
  title={Training data-efficient image transformers \& distillation through attention},
  author={Touvron, Hugo and Cord, Matthieu and Douze, Matthijs and Massa, Francisco and Sablayrolles, Alexandre and J{\'e}gou, Herv{\'e}},
  booktitle={International conference on machine learning},
  pages={10347--10357},
  year={2021},
  organization={PMLR}
}

@inproceedings{wang2021pyramid,
  title={Pyramid vision transformer: A versatile backbone for dense prediction without convolutions},
  author={Wang, Wenhai and Xie, Enze and Li, Xiang and Fan, Deng-Ping and Song, Kaitao and Liang, Ding and Lu, Tong and Luo, Ping and Shao, Ling},
  booktitle={Proceedings of the IEEE/CVF international conference on computer vision},
  pages={568--578},
  year={2021}
}

@inproceedings{yu2022metaformer,
  title={Metaformer is actually what you need for vision},
  author={Yu, Weihao and Luo, Mi and Zhou, Pan and Si, Chenyang and Zhou, Yichen and Wang, Xinchao and Feng, Jiashi and Yan, Shuicheng},
  booktitle={Proceedings of the IEEE/CVF conference on computer vision and pattern recognition},
  pages={10819--10829},
  year={2022}
}

@inproceedings{dosovitskiy2020image,
  title={An Image is Worth 16x16 Words: Transformers for Image Recognition at Scale},
  author={Dosovitskiy, Alexey and Beyer, Lucas and Kolesnikov, Alexander and Weissenborn, Dirk and Zhai, Xiaohua and Unterthiner, Thomas and Dehghani, Mostafa and Minderer, Matthias and Heigold, Georg and Gelly, Sylvain and others},
  booktitle={International conference on learning representations},
  year={2021}
}

@book{kolmogorov1961representation,
  title={On the representation of continuous functions of several variables by superpositions of continuous functions of a smaller number of variables},
  author={Kolmogorov, Andre{\u\i} Nikolaevich},
  year={1961},
  publisher={American Mathematical Society}
}

@inproceedings{deng2009imagenet,
  title={Imagenet: A large-scale hierarchical image database},
  author={Deng, Jia and Dong, Wei and Socher, Richard and Li, Li-Jia and Li, Kai and Fei-Fei, Li},
  booktitle={2009 IEEE conference on computer vision and pattern recognition},
  pages={248--255},
  year={2009},
  organization={IEEE}
}

@inproceedings{loshchilov2017decoupled,
  title={Decoupled Weight Decay Regularization},
  author={Loshchilov, Ilya and Hutter, Frank},
  booktitle={International conference on learning representations},
  year={2019}
}

@inproceedings{he2016deep,
  title={Deep residual learning for image recognition},
  author={He, Kaiming and Zhang, Xiangyu and Ren, Shaoqing and Sun, Jian},
  booktitle={Proceedings of the IEEE conference on computer vision and pattern recognition},
  pages={770--778},
  year={2016}
}

@article{touvron2022resmlp,
  title={Resmlp: Feedforward networks for image classification with data-efficient training},
  author={Touvron, Hugo and Bojanowski, Piotr and Caron, Mathilde and Cord, Matthieu and El-Nouby, Alaaeldin and Grave, Edouard and Izacard, Gautier and Joulin, Armand and Synnaeve, Gabriel and Verbeek, Jakob and others},
  journal={IEEE transactions on pattern analysis and machine intelligence},
  volume={45},
  number={4},
  pages={5314--5321},
  year={2022},
  publisher={IEEE}
}

@article{liu2024kan2,
  title={Kan 2.0: Kolmogorov-arnold networks meet science},
  author={Liu, Ziming and Ma, Pingchuan and Wang, Yixuan and Matusik, Wojciech and Tegmark, Max},
  journal={arXiv preprint arXiv:2408.10205},
  year={2024}
}

@inproceedings{yu2025exploring,
  title={Exploring Kolmogorov-Arnold networks for realistic image sharpness assessment},
  author={Yu, Shaode and Chen, Ze and Yang, Zhimu and Gu, Jiacheng and Feng, Bizu and Sun, Qiurui},
  booktitle={ICASSP 2025-2025 IEEE international conference on acoustics, speech and signal processing (ICASSP)},
  pages={1--5},
  year={2025},
  organization={IEEE}
}

@inproceedings{bastings2020elephant,
  title={The elephant in the interpretability room: Why use attention as explanation when we have saliency methods?},
  author={Bastings, Jasmijn and Filippova, Katja},
  booktitle={Proceedings of the third blackboxNLP workshop on analyzing and interpreting neural networks for NLP},
  pages={149--155},
  year={2020}
}

@article{yang2025medkan,
  title={Medkan: An advanced kolmogorov-arnold network for medical image classification},
  author={Yang, Zhuoqin and Zhang, Jiansong and Luo, Xiaoling and Lu, Zheng and Shen, Linlin},
  journal={arXiv preprint arXiv:2502.18416},
  year={2025}
}

@article{yang2024activation,
  title={Activation space selectable kolmogorov-arnold networks},
  author={Yang, Zhuoqin and Zhang, Jiansong and Luo, Xiaoling and Lu, Zheng and Shen, Linlin},
  journal={arXiv preprint arXiv:2408.08338},
  year={2024}
}

\end{document}